\documentclass[futureinternet,article,accept,pdftex,moreauthors]{Definitions/mdpi} 

\firstpage{1} 
\pubvolume{1}
\issuenum{1}
\articlenumber{0}
\pubyear{2024}
\copyrightyear{2024}
\externaleditor{{Academic Editor:  Stefano Rinaldi and Alan Oliveira De Sá }}  
\datereceived{20 July 2024} 
 \daterevised{12 August 2024 } 
\dateaccepted{13 August 2024 } 
\datepublished{} 
\hreflink{https://doi.org/}

\usepackage[linesnumbered,ruled,vlined]{algorithm2e}
\usepackage[normalem]{ulem}
\useunder{\uline}{\ul}{}
\usepackage{xcolor}
\usepackage{array}
\usepackage{bm}
\usepackage{bbding}
\newcolumntype{Y}{>{\centering\arraybackslash}X}

\Title{Dynamic Graph Representation Learning for Passenger Behavior~Prediction}

\TitleCitation{Dynamic Graph Representation Learning for Passenger Behavior Prediction}

\Author{{Mingxuan} 
 Xie $^{1}$\orcidA{}, Tao Zou $^{1}$, Junchen Ye $^{2}$*, Bowen Du $^{1,2,3}$ and Runhe Huang $^{4}$}

\AuthorNames{Mingxuan Xie, Tao Zou, Junchen Ye, Bowen Du and Runhe Huang}

\AuthorCitation{Xie, M.; Zou, T.; Ye, J.; Du, B; Huang, R.}

\address{%

$^{1}$ \quad CCSE Lab, Beihang University, Beijing, 100191, China; xie\_mingxuan@buaa.edu.cn (M.X.); zoutao@buaa.edu.cn (T.Z.);  dubowen@buaa.edu.cn (B.D.) \\  
$^{2}$ \quad  School of Transportation Science and Engineering, Beihang University, Beijing, 100191, China\\
$^{3}$ \quad  Zhongguancun Lab, Beijing, 100190, China\\
$^{4}$ \quad Faculty of Computer and Information Science, Hosei University, Tokyo, 102-8160, Japan; rhuang@hosei.ac.jp}

\corres{Correspondence: junchenye@buaa.edu.cn}

\abstract{Passenger behavior prediction aims to track passenger travel patterns through historical boarding and alighting data, enabling the analysis of urban station passenger flow and timely risk management. This is crucial for smart city development and public transportation planning. Existing research primarily relies on statistical methods and sequential models to learn from individual historical interactions, which ignores the correlations between passengers and stations. To address these issues, this paper proposes DyGPP, which leverages dynamic graphs to capture the intricate evolution of passenger behavior. First, we formalize passengers and stations as heterogeneous vertices in a dynamic graph, with connections between vertices representing interactions between passengers and stations. Then, we sample the historical interaction sequences for passengers and stations separately. We capture the temporal patterns from individual sequences and correlate the temporal behavior between the two sequences. 
Finally, we use an MLP-based encoder to learn the temporal patterns in the interactions and generate real-time representations of passengers and stations. Experiments on real-world datasets confirmed that DyGPP outperformed current models in the behavior prediction task, demonstrating the superiority of our model.}


\begin{document}


\section{Introduction}
Nowadays, with the rapid development of transportation technology, subways play a significant role in alleviating traffic pressure in large cities and improving the efficiency of residents' travel, thereby accumulating a large amount of passenger travel records. In intelligent transportation city systems, managing the risk of the passenger flow at stations is crucial for enhancing urban safety and maintaining the stability of infrastructure operations. Passenger behavior prediction (PP) models use passengers' historical boarding and alighting information to reflect their travel habits and preferences. The results can be applied to various downstream tasks, such as analyzing future passenger flow trends at stations based on group travel behavior or managing risks at key urban stations based on crowd travel patterns.

The traditional PP models initially relied on statistical methods and conventional time series prediction analysis. Recently, the application of deep learning techniques to PP tasks has demonstrated higher prediction accuracy and better user learning capabilities. The diffusion convolutional gated recurrent unit (DCGRU) model \cite{jia2020urban} first generates subgraphs for traffic conditions, learns subgraph messages through a deep convolutional neural network (DCNN), and then captures user behavior habits through a gate recurrent unit (GRU). A multiple temporal units neural network (MTUNN) \cite{tsai2009neural} uses multiple time units to process information of different time lengths, while a PENN integrates different structural information using various models and then makes the final prediction. A long short-term memory neural network (LSTM\_NN) \cite{liu2020impacts} models the long-term patterns of passengers by applying LSTM and also explores the significant impact of weather on passenger flow. Additionally, many traditional methods \cite{lijuan2020neural, 8543497, gu2022short} predict temporal features using an autoregressive integrated moving average model (ARIMA), analyze passenger flow with a support vector machine (SVM), or forecast short-term changes using a CNN or other methods.


However, the current works overlook three important features of PP tasks:

Firstly, passenger travel records are continuously growing. Current PP models typically only model the existing historical interaction sequences of passengers to make single-step predictions and evaluations for the next moment. In the real world, passenger travel records are continuously increasing, so prediction tasks should incorporate the latest interaction information in a timely manner, and predictions should be multi-step, with the model being promptly updated based on changes to travel records. Sequence-based solutions in PP models cannot adequately adapt to real-life applications.

Secondly, passenger travel behavior encompasses both long-term and short-term characteristics. Human behavior exhibits periodicity, showing regular patterns over the long term but potentially undergoing abrupt changes in the short term. For example, a passenger might typically commute between home and work from Monday to Friday but stay home or visit a shopping center on weekends. Long-term periodic patterns reveal inherent characteristics of passengers, such as identifying them as workers based on their regular commute between work and home. Short-term variability, however, can be influenced by factors such as weather conditions and the passenger's mood. The existing methods often capture the long-term characteristics of passengers but struggle to effectively extract short-term variability features.

Thirdly, passenger travel behavior involves complex interactions between passengers and stations. A passenger may have entered and exited multiple different stations before a certain time, and a station may have had multiple passengers enter and exit before that time. Therefore, the relationship between a passenger and a station is intricate and complex. Relying solely on statistical methods to determine the probability of a passenger interacting with a station at a given time can overlook the influence of passenger density at the station on the passenger's destination choice.

\begin{figure}[H]
\includegraphics[width=13.5 cm]{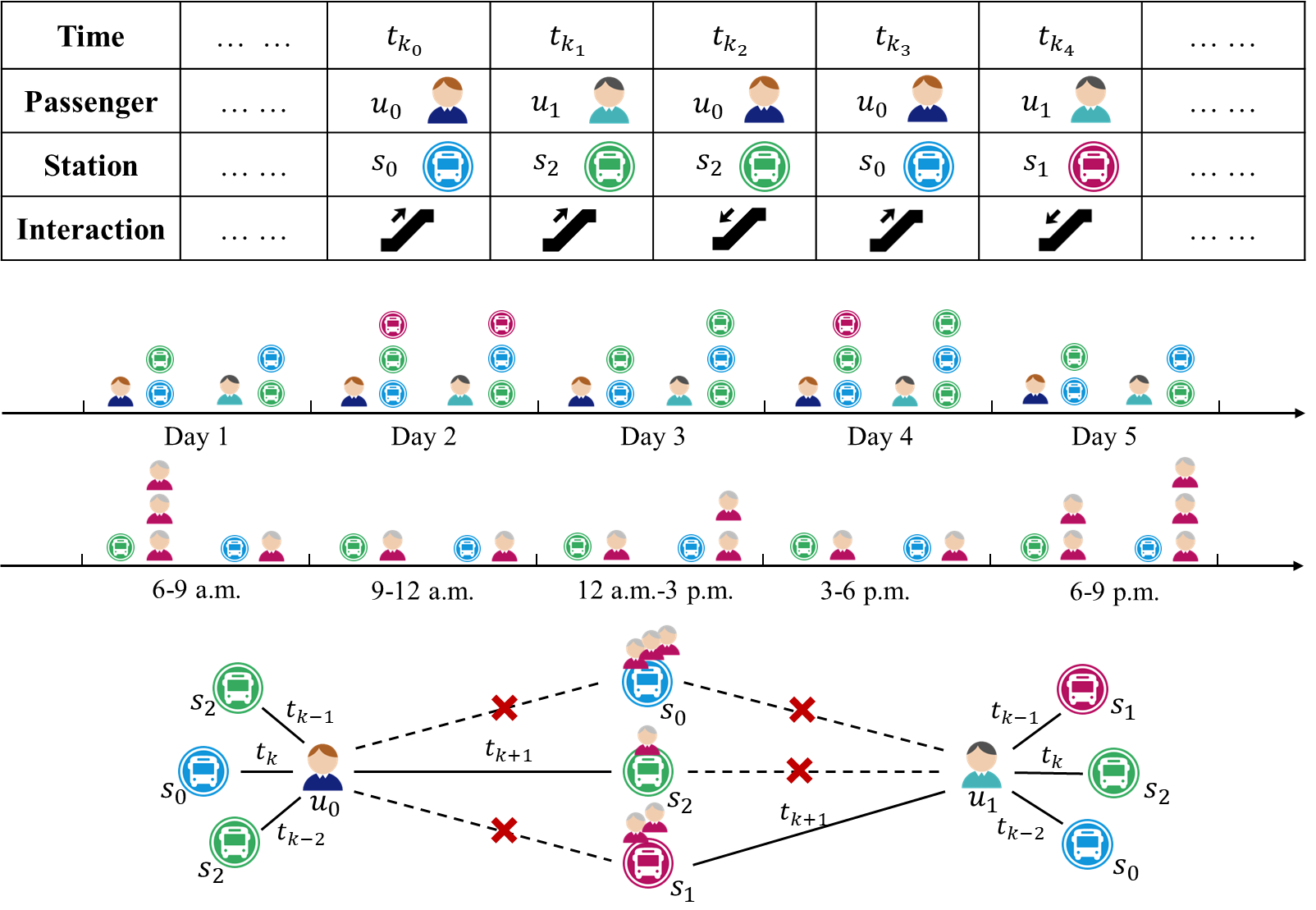}
\caption{Passenger behavior prediction 
with a dynamic graph to trace passengers' travel patterns. This task has three dynamics: (1) travel records are infinitely growing; (2) periodic patterns over long intervals and abrupt changes over short intervals. (3) evolving relationships between passengers and stations}
\label{fig1}
\end{figure}

To address these issues, a new method called \textbf{{dy}
}namic \textbf{{g}}raph representation learning for \textbf{{p}}assenger behavior \textbf{{p}}rediction (DyGPP) is proposed as shown in Figure \ref{fig1}. This method aims to track the dynamic travel behavior characteristics of passengers using continuous-time dynamic graphs (CTDG). In this dynamic graph, passengers and stations are represented as heterogeneous nodes, and the entire graph can be viewed as a bipartite graph. The goal is to predict the connection between specific passenger nodes and station nodes at particular times, indicating whether the passenger will interact with the corresponding station at a certain time. This novel design can adapt to the continuously growing interaction records between passengers and stations, while modeling passengers' long-term periodic features and short-term abrupt features using their historical interaction sequences. Experiments on the Beijing subway travel dataset were conducted, and the results demonstrated the effectiveness of our method. 

The contributions can be summarized as follows:
\begin{itemize}
    \item A continuous-time dynamic graph model is proposed for the passenger behavior prediction problem. Unlike previous sequential models, DyGPP treats fine-grained timestamped events as a dynamic graph and learns temporal patterns for both passengers and stations from their historical sampling sequences.  
    \item A temporal pattern encoding module is designed to simultaneously capture the individual temporal patterns for passengers and stations and correlate temporal behaviors between passengers and stations.
    \item Experiments on real-world datasets demonstrated the effectiveness of our method. Compared to the baseline models, our approach achieved higher average precision and AUC values, demonstrating its superiority.
\end{itemize}

\section{Related Works}
\subsection{Passenger Behavior Prediction}
Passenger behavior prediction has garnered significant attention in urban data analysis. Early research \cite{mcfadden1974measurement} proposed a multidimensional analysis of travel behavior to understand people's travel demands. In recent years, travel behavior analysis has been commonly used in urban function perception and smart transportation development. In machine learning methods, neural network models have been effective in solving time series prediction problems \cite{jia2020urban}. Tsai et~al. \cite{tsai2009neural} considered temporal features in neural network models to predict station-based passenger flow. Wei and Chen \cite{wei2012forecasting} used passenger flow data from the six time steps before the prediction period as input, combining empirical mode decomposition and NN models to predict passenger flow. Menon and Lee \cite{menon2017predicting} combined a non-homogeneous Poisson process with a single-layer neural network model to predict short-term public transit passenger flow. Zhai et al. \cite{zhai2020novel} proposed a new hierarchical hybrid model for short-term passenger flow prediction, based on time series models, deep belief networks, and an improved incremental extreme learning machine. Liu et al. \cite{liu2020impacts} applied LSTM networks to develop an hourly subway passenger flow prediction model. Yao et~al. \cite{lijuan2020neural} used convolutional neural networks to predict short-term passenger flow during special events. The work in ~\cite{8543497} proposed a short-term passenger flow prediction method for subways based on ARIMA, incorporating weather conditions. Gu et al. \cite{gu2022short} proposed a trajectory sequence encoding method and applied it to an interpretable StTP model with adaptive location awareness. Hao et al. \cite{hao2019sequence} proposed an end-to-end framework for multi-step prediction, using an attention mechanism to predict the number of passengers alighting at all stations. Please see Table \ref{tab4} for a review summary.

\begin{table}[H]
\caption{Summary of related works in passenger behavior prediction.} \label{tab4}
\begin{tabularx}{\textwidth}{cccl}
\toprule
\multicolumn{1}{c}{\textbf{Ref}} & \multicolumn{1}{c}{\begin{tabular}[c]{@{}c@{}}\textbf{Time Series}\\ \textbf{Prediction}\end{tabular}} & \multicolumn{1}{c}{\begin{tabular}[c]{@{}c@{}}\textbf{Passenger Flow}\\ \textbf{Prediction}\end{tabular}} & \multicolumn{1}{c}{\textbf{Solution}}                          \\
\midrule
\cite{jia2020urban}                        & \Checkmark                  & \XSolidBrush                   & Neural network models                                 \\
\cite{tsai2009neural}                        & \XSolidBrush                & \Checkmark                     & Neural network models with temporal features          \\
\cite{liu2020impacts}                        & \XSolidBrush                & \Checkmark                     & LSTM models                                           \\
\cite{lijuan2020neural}                        & \XSolidBrush                & \Checkmark                     & CNN models                                            \\
\cite{8543497}                        & \Checkmark                  & \Checkmark                     & ARIMA model with weather conditions                   \\
\cite{gu2022short}                        & \XSolidBrush                & \XSolidBrush                   & StTP model with adaptive location awareness           \\
\cite{wei2012forecasting}                        & \XSolidBrush                & \Checkmark                     & NN models with six time steps before prediction       \\
\cite{menon2017predicting}                        & \XSolidBrush                & \Checkmark                     & Non-homogeneous Poisson process                       \\
\cite{zhai2020novel}                        & \Checkmark                  & \Checkmark                     & Hierarchical hybrid model based on time series models \\
\cite{hao2019sequence}                        & \Checkmark                  & \XSolidBrush                     & End-to-end framework for multi-step prediction \\
\bottomrule
\end{tabularx}
\end{table}

\subsection{Dynamic Graph Learning}
 Dynamic graphs describe entities as nodes and represent their interactions as edges with timestamps. In recent years, extensive research \cite{barros2021survey, kazemi2020representation} on dynamic graph representation learning has helped simulate real-world network scenarios more effectively. Research on dynamic graphs can be divided into two types: one decomposes the changing graph structure at time intervals and takes snapshots at the final moment of decomposition \cite{you2022roland, zhang2023dyted}. The dynamics of the graph are reflected in the changes in these snapshots. However, this method cannot effectively represent all interaction changes in a dynamic graph; it can only show the overall trend. To analyze finer-grained changes in graph structure, some methods add time-based random walks \cite{nguyen2018dynamic, yu2018netwalk} to the graph structure to represent node information, while others preserve temporal information in node features \cite{rossi2020temporal, kumar2019predicting, tolstikhin2021mlp}. Additionally, some approaches use sequence models to capture long-term temporal features in dynamic graph structures \cite{wang2021tcl, yu2023towards}. Jodie \cite{kumar2019predicting} uses an RNN-based encoder to model the evolution of nodes. TGAT \cite{xu2020inductive} extracts and encodes local temporal subgraphs to represent the current state of nodes. TGN \cite{poursafaei2022towards} extends Jodie with a graph convolutional layer to aggregate information from local neighbors and updates node representations with the aggregated temporal information.

\subsection{Time-Based Batch}
When calculating the loss for gradient descent to update weights \cite{ruder2016overview,darken1992learning,bengio2009curriculum}, we can use all the samples to compute the loss; this method is called batch gradient descent (BGD). Alternatively, we can randomly select a single sample to compute the loss and then the gradient; this method is called stochastic gradient descent (SGD). To balance the advantages of BGD and SGD, mini-batch gradient descent (MBGD) was developed. The Adam optimizer
\cite{kingma2014adam} further enhances the MBGD method by introducing first moment and second moment estimates to control the learning rate. However, when it comes to dynamic link prediction problems, the effectiveness of the aforementioned methods may decrease as the batch size increases. Lampert \cite{lampert2024link} proposed dynamic link forecasting, which maintains prediction accuracy by dividing batches using fixed time intervals. This approach enhances the robustness of the model.

\section{Preliminaries}
In this section, we first present some necessary definitions and then formalize the studied problem. The notations used and their meanings are shown in the Table  \ref{tab3}.

\begin{table}[H] 
\caption{Notations and their meanings.\label{tab3}}
\renewcommand{\arraystretch}{1.1} 
\begin{tabularx}{\textwidth}{cL}
\toprule
\textbf{Notation}	& \textbf{Meaning}\\
\midrule
$u_i$         & The $i^{th}$ passenger \\
$s_i$         & The $i^{th}$ station \\
$\mathcal{U}$ & Collections of passengers \\
$\mathcal{S}$ & Collections of stations \\
$l_k$         & The $k^{th}$ label, which is 0 when the passenger gets on the train and 1 when they get off the train  \\
$t_k$         & The $k^{th}$ timestamp  \\
$e_k$         & The $k^{th}$ interaction \\
$\mathcal{G}^t$ & The dynamic graph at time $t$ \\
$x_{u_i}$       & Feature of the passenger $u_i$ \\
$x_{s_i}$       & Feature of the station $s_i$ \\
$\hat{p}_{u,s}^{t}$ & The probability of interaction between passenger $u$ and stations $s$ at time $t$ \\
FC  & Fully connected layer \\
ReLU & Linear rectification function, a type of activation function in neural networks \\
Sigmoid & A common S-shaped function in biology, it also acts as the activation function for neural networks, mapping variables between 0 and 1 \\
\bottomrule
\end{tabularx}
\end{table}

\subsection{Definitions}

Let $\mathcal{U} = \{ u_i \mid i = 1, 2, \ldots, n \}$ and $\mathcal{S} = \{ s_j \mid j = 1, 2, \ldots, m \}$ represent the collections of $n$ passengers and $m$ stations. Specifically, $u_i$ represents the $i$ -th passenger and $s_j$ represents the $j$-th station.

\begin{Definition}
    \textbf{{Passenger Behavior.} 
} The passenger behavior is defined as an interaction $e_k=(u_k, s_k, l_k, t_k)$ where the passenger $u_k$ enters or leaves the subway station $s_k$ at timestamp $t_k$.
\end{Definition}

\begin{Definition}
    \textbf{{Dynamic} Graph.} A dynamic graph can be represented as a monotonically non-decreasing sequence of interactions denoted by $\mathcal{G}^t=\left\{e_1,\cdots,e_k\right\}$, where $0 \textless t_1 \leq \cdots \leq t_k \leq t$. In the graph, each passenger node $u_i$ is associated with a node feature $x_{u_i} \in \mathbb{R}^{d_N}$, and the feature of each station node $x_{s_j} \in \mathbb{R^{d_N}}$ also carries the same meaning. Additionally, we defined corresponding features for the edges $x_{e_k}^{t_k} \in \mathbb{R}^{d_E}$, where ${d_N}$ and ${d_E}$ denote the dimensions of the node feature and the edge feature.
\end{Definition}

\subsection{Problem Formalization.} 

\textbf{{Passenger} 
 Behavior Prediction} with a dynamic graph aims to predict the subway station that passenger $u_i\in\mathcal{U}$ will enter or leave in the future. Specifically, given the passenger node $u_i$, and the historical passenger--station interaction sequence before $t_k$, i.e.,  $\mathcal{G}^{t_k}$, we solve such a problem by predicting whether $u_i$ will pass through station $s_j\in\mathcal{S}$ at timestamp $t_k$. The process can be formalized as below

$$\hat{p}_{u_i,s_j}^{t_k} = f(\mathcal{G}^{t_k})$$
where $\hat{p}_{u_i,s_j}^{t_k}$ is the probability that passenger $u_i$ will pass through the target interaction station $s_j$ at time $t_k$.





\section{Methodology}
\begin{figure}[H]
\includegraphics[width=13.5 cm]{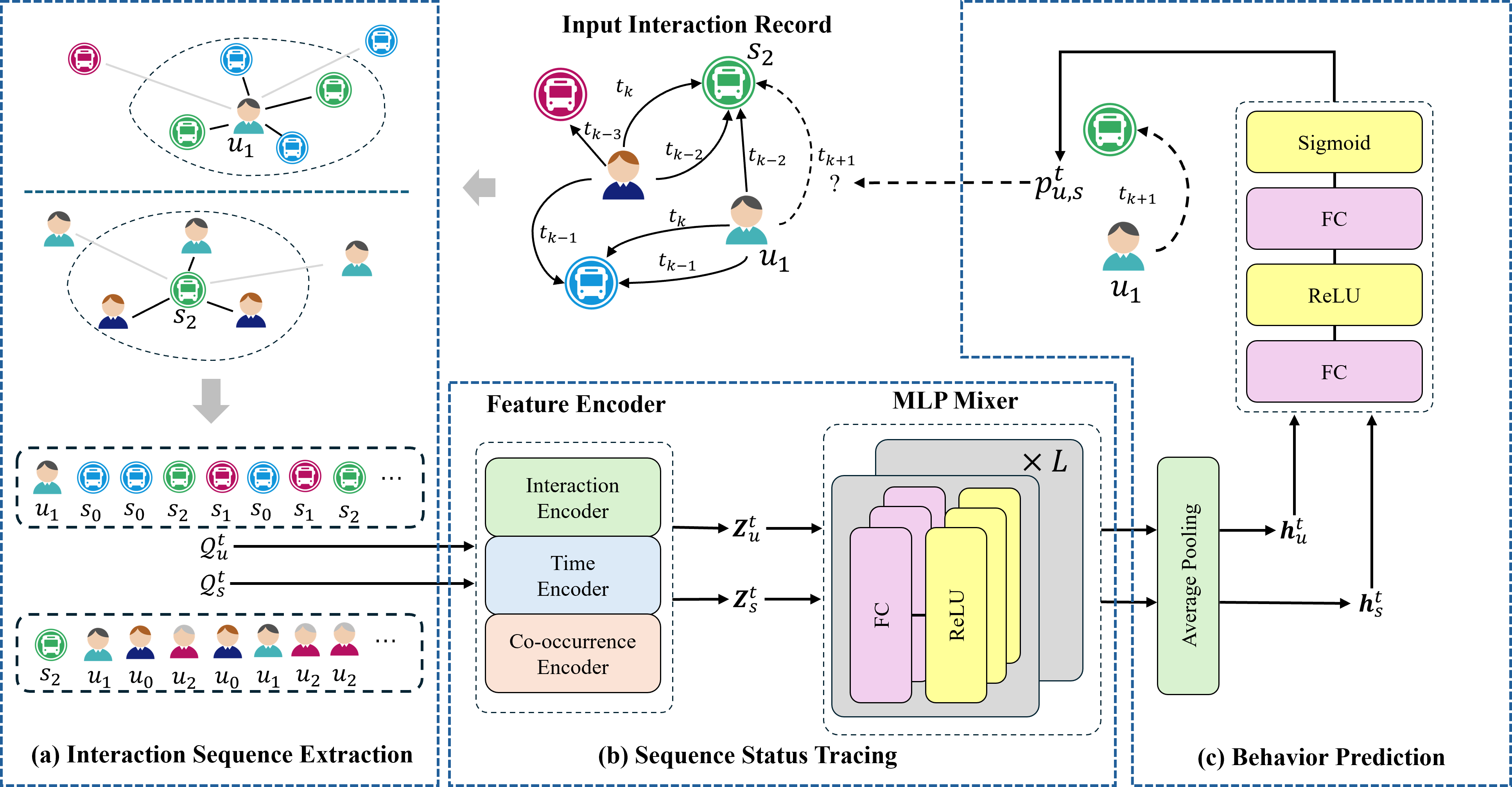}
\caption{Framework of the DyGPP model.}
\label{fig2}
\end{figure}

The overall architecture of our method is illustrated in Figure  \ref{fig2}. The model for learning passenger travel behavior can be divided into three components: temporal sequence construction, dynamic representation learning, and future behavior prediction. Specifically, given a predicted interaction $e_k$ at time $t_k$, we first fetch the recent historical interaction sequence and capture the temporal features of passenger $u_k$ and station $s_k$ via the first component. The second component learns the dynamic behavior pattern for the passenger and passenger flow trend for the station before time $t$. Then, a prediction layer is utilized to estimate the probability that passenger $u_k$ would pass through station $s_k$ based on their dynamic representations. Finally, we present a time-batch algorithm to enhance the model efficiency by creating batches based on time intervals, ensuring consistency with the passenger flow over time.


\subsection{Temporal Sequence Construction}
Passenger travel behavior tends to be regular in traffic systems, particularly for metro systems \cite{cheng2021incorporating, cantelmo2019incorporating}. To capture the temporal patterns, we first sample the historical first-hop neighbors of passengers and stations. Then, we encode the temporal information and interaction behavior between two nodes via different feature encoders.


\subsubsection{Extracting Historical Sequences} 

The sequence of historical neighbors always reflects the temporal patterns of nodes in the dynamic graphs \cite{xu2020inductive, rossi2020temporal, wang2021tcl}. To capture the evolving patterns for passengers and stations, we extract a fixed number of first-order neighbors from an infinitely long historical interaction sequence and construct a corresponding interaction sequence. Formally, this can be expressed as follows: Given a potential interaction $e=(u, s, l, t)$ at time $t$, we extract the corresponding interaction sequences $\mathcal{Q}_u^t = \{(u, s^\prime, l^\prime, t^\prime) \mid t^\prime < t\} \cup \{(u, u, 0, t)\}$ and $\mathcal{Q}_s^t = \{(u^\prime, s, l^\prime, t^\prime) \mid t^\prime < t\} \cup \{(s, s, 0, t)\}$ for the passenger node $u$ and the station node $s$ with label $l$ before time $t$. Specifically, we consider that the state of passengers and stations at time $t$ is related to their characteristics, so we treat the node itself as a neighbor and include it in the model. To accurately represent the current node features, we sort the interactions by time and select the $N$ recently interacted neighbors up to time $t$. For nodes with fewer neighbors, we use a zero-padding technique to expand the historical neighbors, ensuring that each node has $N$ neighbors. This method improves the efficiency of the model without sacrificing accuracy and maintains the stability of model training.

\subsubsection{Feature Encoder}
For each interaction $e=(u,v,l,t)$, we first obtain the embeddings for each sequence with interaction features, time interval information, and temporal pattern behaviors. Then, we generate the fused representations for each passenger and station by aggregating these embeddings. 

\textbf{{Interaction} Encoder.}
Boarding and alighting, as key attributes of passenger--station interactions, are encoded into the interaction features. We use different labels to distinguish between boarding and alighting and write them into the raw features of the edges. Next, we use a replication padding method to expand the raw feature matrix of the edges, which serves as $\mathbf{X}_E$ in the feature fusion.

\textbf{{Time} Encoder.}
Passenger boarding and alighting behaviors usually exhibit strong periodicity and are highly correlated with time. We developed a time encoder to effectively utilize the time features in historical interaction information. Let the time encoding dimension be $d$, the predicted timestamp be $t$, and for passenger $u$'s neighbor $s_i$ interacted at time $t_i$, we use the $\cos{(\cdot)}$ to transform the time interval $\Delta t = t - t_i$ into a high-dimensional vector, and then represent the time interval using learnable parameters, followed by \cite{cong2023we}. The formula is as follows:
\begin{linenomath}
\begin{equation}
\textbf{x}_{s_i, time}^{t} = \sqrt{\frac{1}{d_T}} \left[ \cos{(\omega_1 \Delta t)}, \cos{(\omega_2 \Delta t)}, \ldots, \cos{(\omega_{d_T} \Delta t)} \right],
\end{equation}
\end{linenomath}
where $\textbf{x}_{s_i, time}^{t} \in \mathbb{R}^{d_T}$ and $\bm{\omega} = [\omega_1, \ldots, \omega_{d_T}]^T$ represent the trainable parameters. Hence, the time interval embeddings for the sequence of the passenger and station are denoted as $\bm{X}^t_{u, T}$ and $\bm{X}^t_{s, T}$, respectively.


\textbf{{Temporal} Pattern Encoding.} The existing methods usually calculate the feature information of the passenger sequence and the station sequence separately, ignoring the correlation between the two nodes. We designed a module to consider both separate and cross-temporal patterns in these two sequences. It is worth noting that passenger nodes and station nodes are two different types of node. The historical first-order neighbors of a passenger node are always station nodes, and the historical first-order neighbors of a station node are always passenger nodes. In other words, the historical neighbor set of a passenger node can be denoted as $\mathcal{N}_u^t = \{u\} \cup \{s_1^{t_1}, \ldots, s_n^{t_n} \mid t_n < t\}$, and the station nodes' neighbor set can be denoted as $\mathcal{N}_s^t = \{s\} \cup \{u_1^{t_1}, \ldots, u_n^{t_n} \mid t^{t_n} < t\}$.

On the one hand, to capture the separate temporal patterns, we calculate the interaction frequency of each neighbor in the individual sequence, which aims to learn the long-term temporal behaviors of passengers and stations. On the other hand, we learn the repeated interaction frequency between the passenger and target station in all other sequences. Specifically, we calculate the frequency at which passenger nodes appear in the historical neighbors of station nodes and the frequency at which station nodes appear in the historical neighbors of passenger nodes. The higher the frequency, the stronger the correlation between the two. In other words, if a station node $u$ appears frequently in the historical neighbors of a passenger $\mathcal{N}_u^t$, it is more likely that the passenger will interact with the station at time $t$.

\begin{figure}[H]
\includegraphics[width=7.5 cm]{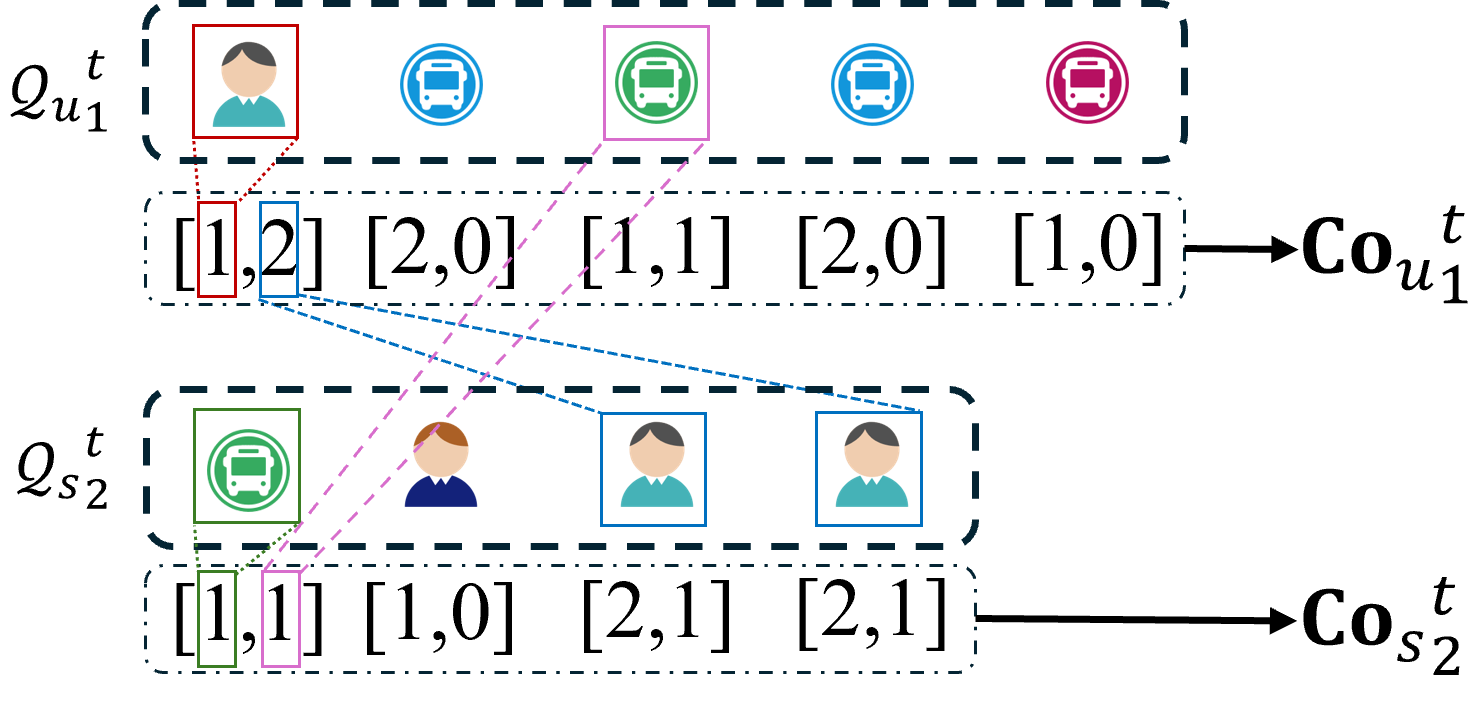}
\caption{Example of co-occurrence features.}
\label{fig3}
\end{figure}

Formally speaking, given the historical interaction sequence $\mathcal{Q}_u^t$ for passengers and $\mathcal{Q}_s^t$ for stations, we calculate the correlation between the two sequences, denoted as 
 $\mathbf{Co_u^t}$ and $\mathbf{Co_s^t}$. For example, let us assume the historical neighbor sequence of $u_1$ as $\mathcal{Q}_{u_1}^t = \{u_1, s_1, s_2, s_1, s_3\}$, and the historical neighbor sequence of $s_2$ as $\mathcal{Q}_{s_2}^t = \{s_2, u_2, u_1, u_1\}$ as shown in Figure ~\ref{fig3}. The frequency at which $u_1$ appears in the $\mathcal{Q}_{u_1}^t, \mathcal{Q}_{s_2}^t$ sequence is $[1,2]$, then the correlation can be denoted as $\mathbf{Co}_{u_1}^t = \{[1,2],[2,0],[1,1],[2,0],[1,0]\}$. Similarly, $\mathbf{Co}_{s_2}^t$ can be denoted as $\mathbf{Co}_{s_2}^t = \{[1,1], [1,0], [2,1], [2,1]\}$. 

Then, we apply a function $f$ to generate the temporally correlated representation by
\begin{linenomath}
\begin{equation}
\mathbf{X}^t_{*,Co} = f \left( \mathbf{Co^t_*[:,0]} \right) + f \left( \mathbf{Co^t_*[:,1]} \right),
\end{equation}
\end{linenomath}
where $*$ represents $u$ or $s$.

\textbf{{Feature} Fusion.} After extracting the sequence information, we obtain the historical interactions that occurred before time $t$ for both passengers and stations. These interactions can be used to represent the current node features. We further introduce learnable features $\mathbf{X}^U_{N} \in \mathbb{R}^{n \times d_N}, \mathbf{X}^S_{N} \in \mathbb{R}^{m \times d_N}$ for passenger nodes and station nodes to represent their inherent attributes, which will be gradually learned during the model training process. Hence, we denote $\mathbf{X}_{u, N}^t, \mathbf{X}_{s, N}^t$ as the learnable node feature sequences for passenger $u$'s neighbors and station $s$'s neighbors. Additionally, for historical interaction information $e$, the boarding and alighting attributes of passengers are important features for representing this interaction information. Therefore, we represent the fused features as follows:


\begin{equation}
\mathbf{Z}_*^t = Proj \left( \mathbf{X}_{*, N}^t ||  \mathbf{X}_{*, E}^t || \mathbf{X}_{*, T}^t || \mathbf{X}_{*, Co}^t \right),
\end{equation}
where $*$ denotes $u$ or $s$. $Proj(\cdot)$ represents the projection mapping function, which is implemented using a linear layer in this model. This is used to integrate the temporal feature information in the sequence, thereby generating the embedding $\mathbf{Z}_*^t \in \mathbb{R}^{N \times d_E}$ of the corresponding node at time $t$. 

\subsection{Dynamic Representation Learning}
The passenger representations and station representations processed by the projection function have the same dimensions. This reflects the historical boarding and alighting behaviors of passengers, as well as their state (whether they are on or off the train) during the current interaction, over time. Additionally, we incorporate the inherent attributes of the station and the passenger flow changes over time. After obtaining the serialized representations, our goal is to use these representations to track changes in the boarding and alighting states of passengers.

\textbf{{Temporal} Information Aggregation.}
To achieve this goal, we use an MLP block composed of stacked feedforward neural networks (FFN) and activation functions. Considering the issue of potential overfitting due to the small number of boarding and alighting events for a single passenger, we use dropout to address this problem. The processed representations $\mathbf{Z}_*^t$ are then passed through the stacked MLP modules. The entire process is as follows:
\begin{linenomath}
\begin{equation}
\mathbf{r}^l \sim \text{Bernoulli}(p),
\end{equation}
\begin{equation}
\text{Dropout}(\mathbf{h}^l) = \hat{\mathbf{h}}^l = \mathbf{h}^l \odot \mathbf{r}^l,
\end{equation}
\begin{equation}
\text{FFN} (\mathbf{Z_*^t}, \mathbf{W}, \mathbf{b}) = \text{Dropout}(\text{ReLU}(\mathbf{Z_*^t}\mathbf{W} + \mathbf{b})),
\end{equation}
\begin{equation}
\textbf{H}_*^{t} = \text{FFN}(\mathbf{Z_*^t}, \mathbf{W}, \mathbf{b}).
\end{equation}
\end{linenomath}
where $\mathbf{W} \in \mathbb{R}^{d_E \times d_E}$, $\mathbf{b} \in \mathbb{R}^{d_E}$. After processing by the MLP model, the input features $\mathbf{Z}_u^{t,0}$, $\mathbf{Z}_s^{t,0}$ are transformed into output features $\mathbf{H}_u^{t}$ and $\mathbf{H}_s^{t}$. This transformed feature can be used for subsequent downstream prediction tasks.

\textbf{{Node} Embedding Representation}
The representations of passenger nodes and station nodes at time $t$ can be denoted as $\mathbf{h}_p^t$ and $\mathbf{h}_s^t$. These are obtained by averaging the representations $\mathbf{H}_*^t$ and then passing them through a linear output layer. The specific process is as follows:
\begin{linenomath}
\begin{equation}
\mathbf{h}_u^t = \left(\frac{1}{N}\sum_{k=1}^N \mathbf{H}_u^t [k, :]\right) \mathbf{W}_{out} + \mathbf{b}_{out},
\end{equation}
\begin{equation}
\mathbf{h}_s^t = \left(\frac{1}{N}\sum_{k=1}^N \mathbf{H}_s^t [k, :]\right) \mathbf{W}_{out} + \mathbf{b}_{out}.
\end{equation}
\end{linenomath}
where $\mathbf{W}_{out} \in \mathbb{R}^{d_E \times d_{out}}$ and $\mathbf{b}_{out} \in \mathbb{R}^{d_{out}}$ are trainable weights, and $\mathbb{R}^{d_{out}}$ represents the output embedding dimension.

\subsection{Future Behavior Prediction}
We consider the boarding or alighting relationship between passengers and stations as the creation of an edge in a dynamic graph. Therefore, the problem can be transformed into a link prediction problem in a dynamic graph. The representations of passengers and stations are fed into a fully connected network to calculate the probability of a link being formed between them at time $t$. 
\begin{linenomath}
\begin{equation}
\mathbf{Z}^t = \mathbf{h}_u^t || \mathbf{h}_s^t,
\end{equation}
\begin{equation}
p_{u,s}^t = (\text{Relu}(\mathbf{Z}^t \mathbf{W}_1) + \mathbf{b}_1)\mathbf{W}_2 + \mathbf{b}_2.
\end{equation}
\end{linenomath} 
The dimension of $\mathbf{Z}$ is $2 \times d_{out}$, the dimension of $\mathbf{W}_1$ and $\mathbf{b}_1$ are $2 \times d_{out}$, while $\mathbf{W}_2 \in \mathbb{R}^{1}$, $\mathbf{b}_2 \in \mathbb{R}^1$. $p_t$ represents the probability of a new link being formed between the passenger node and the station node at time $t$.

\subsection{Parameter Optimization Algorithm}
In our approach, the model training process begins with the construction of a neighbor sampler. For each training iteration, we first set a random seed to randomize the parameters and then load the model and start training. For each batch, we first calculate the most recent first-order neighbors for all passengers and stations in the batch. Next, we pad the extended first-order interaction sequences to generate historical interaction sequences of the same length. We then calculate the correlation of the sequences and use an MLP to capture the temporal characteristics between the sequences. Finally, we estimate the probability of creating new links between passengers and stations and update the parameters using the gradient descent method. The complete algorithm workflow is outlined in Algorithm \ref{alg_training_process}.


\begin{algorithm}[h]
    \SetKwComment{Comment}{/* }{ */}
    \SetKwInOut{Input}{Input}
    \SetKwInOut{Output}{Output}
    \caption{Training 
 process of DyGPP}
 \label{alg_training_process}
    \Input{Collection of passenger $\mathcal{U}$ and collection of stations $\mathcal{S}$, a sequence of historical interaction events $\mathcal{G}^t = \{e_1, \ldots, e_k\}$ with $0 \le t_1 \le \ldots \le t_k \le t$, maximum number of training epochs $MaxEpoch$;}
    \Output{The model parameters $\Theta$ after training;}
    Initialize the parameters in DyGPP with random weights $\Theta$\ and  set $Epochs \gets 1$\;
    \While{not converged and $Epochs \le MaxEpoch$} {
        \For{batch $B_b \in \mathcal{B}$} {
            $\mathcal{N}_{u_i}^t \gets$ Sample latest one-hop neighbors of $u_i^t \in \mathcal{U}$\;
            $\mathcal{N}_{s_i}^t \gets$ Sample latest one-hop neighbors of $s_i^t \in \mathcal{S}$\;
            $\mathcal{S}_{u,i}^t \gets$ Create the interactions of passengers' neighbor set $\mathcal{N}_{u_i}^t$\;
            $\mathcal{Q}_{s_i}^t \gets$ Create the interactions of stations' neighbor set $\mathcal{U}_{s_i}^t$\;
            $\mathcal{Q}_{u_i,pad}^t, \mathcal{Q}_{s,i,pad}^t \gets$ Pad sequences to a regular size\;
            $\mathbf{x}_{u_i,Co}^t, \mathbf{x}_{s_i,Co}^t \gets$ Calculate correlations between $\mathcal{Q}_{u_i,pad}^t$ and $\mathcal{Q}_{s,i,pad}^t$\;
            $\mathbf{x}_{*_i,N}^t, \mathbf{x}_{*_i,E}^t, \mathbf{x}_{*_i,T}^t \gets$ Get raw feature embedding from the origin input\;
            $\mathbf{X}_{*_i}^t \gets$ Generate co-occurrence feature through projection layer\;
            $\mathbf{Z}_{*,i}^t \gets$ Encode passenger embedding and station embedding via MLP network\;
            $\hat{p}_{u,s}^t \gets$ Compute the probabilities by fusing sequence embedding $\mathbf{Z}_{u,i}^t, \mathbf{Z}_{s,i}^t$
            Optimize the model parameters $\theta$ by back propagation\;
        }
        $Epochs \gets Epochs + 1$\;
    }
\end{algorithm}

Batch partitioning is an important method to enhance the parallel training capability of a model and shorten the training time. It should be noted that our approach to batch partitioning differs from commonly used methods. Traditional methods usually determine the batch size based on the number of nodes in each batch, which often works well for time-insensitive data. In our model, the interactions between passengers and stations are strongly time-correlated. Factors such as commuting times and daily traffic flow cycles cause interactions to often concentrate during specific hours of the day. To ensure that the time intervals between data points in each batch are consistent, we adopted a method of partitioning batches based on time intervals.

Formally, given a dynamic graph $\mathcal{G}^t$, batches can be partitioned as $\mathcal{G}^t = \{\mathcal{B}_1, \ldots, \mathcal{B}_n\}$, $\mathcal{B}_i = \{ (u_{i,1},s_{i,1},l_{i,1},t_{i,1}), \ldots, (u_{i,k},s_{i,k},l_{i,k},t_{i,k}) \mid t_{i,1} \le \ldots \le t_{i,k}, t_{i,k} - t_{i,1} \le \delta t \}$, where $\delta t$ is a hyperparameter partitioned to represent the maximum time interval of interactions within a batch. This means that the length of each batch may vary. During periods of high passenger flow, a batch may contain a large amount of interaction information, while during periods of low passenger flow, a batch may contain very little interaction information. This method of batch partitioning ensures the stability of the number of batches across different data scales and provides good temporal information within batches. It also prevents the loss of temporal information, which could occur when multiple batches are within the same time period due to a large number of interactions.

\section{Experiments}
\subsection{Dataset}
In our experiments, we used two real-world datasets, both were collected from the Beijing subway system. The datasets contain all passenger entry and exit data from  1 June 2017 to  9 September 2017. The statistical information of the data is shown in Table \ref{tab1}. 
\begin{table}[H] 
\caption{Statistics of all datasets.\label{tab1}}
\begin{tabularx}{\textwidth}{CCCCC}
\toprule
\textbf{Dataset}	& \textbf{Passenger}	& \textbf{Station}  & \textbf{Interaction} & \textbf{Interaction Per Passenger}\\
\midrule
BJSubway-40K  & 459			        & 392               & 44,852 & 97\\
BJSubway-1M   & 7491			        & 392               & 958,076 & 127\\
\bottomrule
\end{tabularx}
\end{table}

\begin{figure}[H]
\includegraphics[width=13.5 cm]{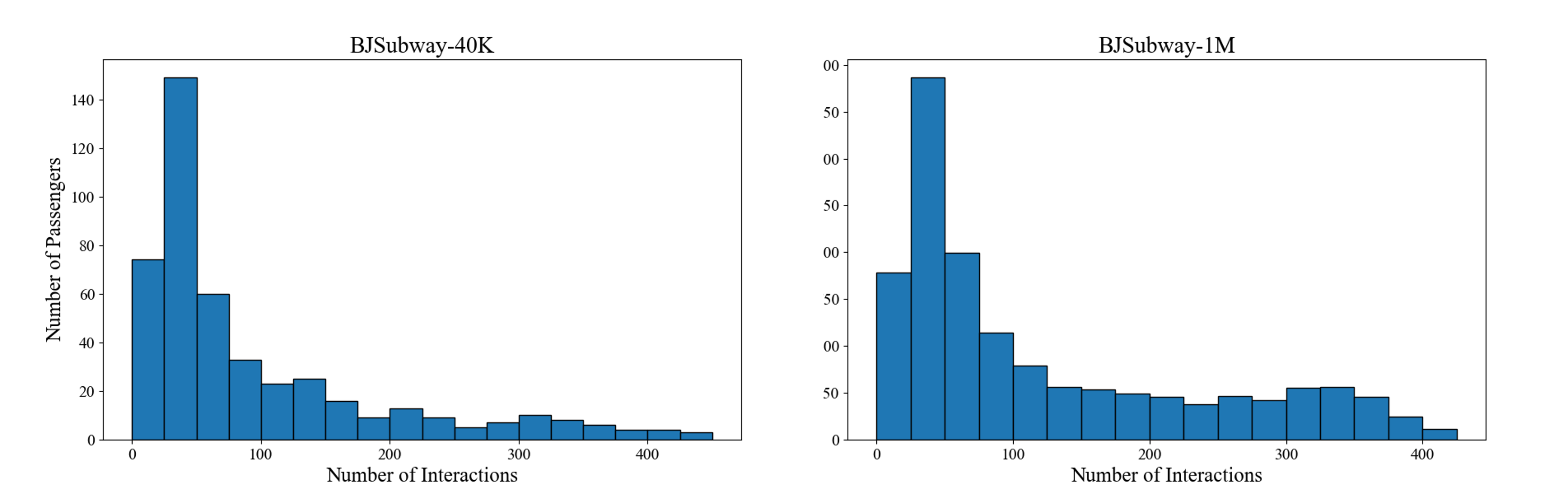}
\caption{Distribution of Interaction Frequencies Between Passengers and Stations.}
\label{fig4}
\end{figure}

We analyzed the distribution of interaction frequencies between different passengers and stations in the two datasets, as shown in Figure  \ref{fig4}. The results indicated that most passengers interacted with stations (including boarding and alighting) fewer than $200$ times. In the 40K dataset, the median was $54$, the Q1 (first quartile) was $30$, and the Q3 (third quartile) was $127$. In the 1M dataset, the median was $72$, the Q1 was $34$, and the Q3 was $202$.

\vspace{-9pt}

\subsection{Baselines}
In our experiments, we compared our model with seven other commonly used models. These models encompass a variety of techniques and implementation methods, including statistical methods (i.e., TOP and Personal TOP), sequence model methods (i.e., LSTM and GRU), graph convolutional methods (i.e., TGAT), memory network models (i.e., TGN), and MLP-based methods (i.e., GraphMixer).

\textbf{{TOP}.} 
The TOP method \cite{DBLP:conf/cikm/Karypis01} is a statistical method based on the travel patterns of all users. First, the training data are fed into the model for statistical analysis. Then, for each station, two sequences are generated: one for all alighting stations when the station is the boarding station, and one for all boarding stations when the station is the alighting station. After completing the statistics, the corresponding station is determined based on the passenger's last travel status, and the station with the highest occurrence frequency in the corresponding sequence is selected as the predicted station.

\textbf{{Personal} TOP.}
The personal TOP \cite{DBLP:conf/cikm/Karypis01} method is similar to the TOP method but generates unique sequences for each specific passenger for boarding and alighting stations. During testing, for passengers with existing sequence records, the predicted station is directly generated based on their unique sequences. For passengers without sequence records, the predicted station is determined based on the overall sequence.

\textbf{{LSTM}.}
LSTM \cite{hochreiter1997long} is a commonly used method for processing sequential data. Compared to traditional RNN models, it introduces input gates, forget gates, output gates, and a cell state, which help prevent the problems of vanishing or exploding gradients. This allows LSTM to better handle long-term dependencies in sequences.

\textbf{{GRU}.}
GRU \cite{cho2014learning} can be seen as a variant of LSTM. It uses a reset gate to determine how to combine new input information with previous memory, and an update gate to define the amount of previous memory to retain for the current time step. GRU is capable of preserving information from long sequences, without clearing it over time or removing it if it is unrelated to the prediction.

\textbf{TGAT.}
TGAT\cite{xu2020inductive} employs a self-attention mechanism to generate corresponding representations by aggregating the temporal topological features of neighbors for each passenger node and station node. Additionally, it uses a time encoder to capture temporal characteristics.

\textbf{TGN.}
TGN\cite{rossi2020temporal} maintains an updated memory node for each passenger node and station node and updates its memory when the node is observed. This is achieved through interaction with a message function, a message aggregator, and a memory updater. This model generates temporal embeddings for both passenger nodes and station nodes.

\textbf{GraphMixer}
GraphMixer\cite{cong2023we} demonstrates that a fixed-time encoding function outperforms the trainable version. It incorporates this fixed function into a link encoder based on MLP-Mixer to learn from temporal links. A node encoder using neighbor mean-pooling is implemented to summarize node features.

\subsection{Evaluation Metrics}
Regarding evaluation metrics, we used the average precision (AP) and area under the receiver operating characteristic curve (AUC) to assess the performance of all methods in predicting binary future connections between passengers and stations. AP measures the proportion of correct predictions among all responses, while AUC reflects the model's ability to distinguish between positive and negative responses.
\begin{linenomath}
\begin{equation}
\text{AP} = \frac{1}{n} \sum_{k=1}^n \frac{\text{TP}_k}{\text{TP}_k + \text{FP}_k},
\end{equation}
\end{linenomath}
where $\text{TP}_k$ represents the cases in the $k$-th prediction where the predicted value matches the true value and is positive (i.e., true positive); FP represents the cases in the $k$-th prediction where the predicted value is positive but the true value is negative (i.e., false positive).

Additionally, we introduced an inductive sampling method to test the accuracy of predicting interactions with stations that passengers have never encountered before. This sampling method involved hiding the station nodes to be used for inductive testing during the initial training phase, and then providing these test station nodes during the testing phase. The proportions used for the training, validation, and test sets were 0.7, 0.15, and 0.15, respectively.

\subsection{Implementation Details}
To ensure consistency and effectiveness in the comparison, the parameter settings for the baseline models were as shown in Table ~\ref{tab2}. We applied the Adam optimizer, which uses the Adam algorithm for first-order gradient-based optimization of stochastic objective functions \cite{kingma2014adam}, and set the early stopping patience to $20$. For all datasets, we set the maximum time interval in a batch to $1000$. For the TOP and personal TOP methods, we constructed an overall station counter and a personal station counter, respectively, to calculate the probability of interactions between passengers and stations. In the LSTM and GRU methods, we input all station interaction records of the passenger to be predicted as a sequence, divide the batch according to time intervals for training, and finally obtained the interaction probability with the target station. In graph-based methods (i.e., TGAT, TGN, GraphMixer, and DyGPP), we first constructed a dynamic graph of all user--station interactions. Then, we divided the node change sequences according to the set time intervals and finally proceeded with the subsequent tasks. For all models, the dimensions of the passenger nodes and station nodes were set to $172$, and the dimensions of the time encoding nodes were set to $100$. The other model settings were consistent with those described in the original articles. The experiments were conducted on an Ubuntu machine featuring two Intel(R) Xeon(R) Gold 6130 CPUs @ 2.10GHz with 16 physical cores. The GPU device utilized was an NVIDIA Tesla T4 with 15GB memory.

\begin{table}[H] 
\caption{Parameters of the experiment.\label{tab2}}
\begin{tabularx}{\textwidth}{LL}
\toprule
\textbf{Parameter}	& \textbf{Value}\\
\midrule
Time gap for batch & 1000s\\
Number of neighbors & 20\\
Neighbor sampling strategy & Most recent neighbor\\
Negative sample strategy & Random\\
Time scaling factor & 0.000001\\
Number of heads & 2\\
Number of encoder model layers & 1\\
Channel embedding dimension & 50\\
Learning rate & 0.0001\\
Max input sequence & 32\\
\bottomrule
\end{tabularx}
\end{table}

\subsection{Performance Comparison}
Table  \ref{tab_all_result} presents the prediction results of whether a passenger interacted with the next station in the two datasets, compared with seven other methods. The results shown in the table are averages obtained after five independent training rounds. The best results are highlighted in bold, and the second-best results are marked with an underline.

\begin{table}[H]
\caption{AP and AUC for passenger behavior prediction.}
\label{tab_all_result}
\begin{tabularx}{\textwidth}{YYYYY}
\toprule
\textbf{Datasets}     & \multicolumn{2}{c}{\textbf{BJSubway-40K}}              & \multicolumn{2}{c}{\textbf{BJSubway-1M}} \\
\midrule
\textbf{Metrics}      & \textbf{AP}                       & \textbf{AUC}                      & \textbf{AP}                  & \textbf{AUC}                \\
\midrule
TOP          & 0.7880 ± 0.0000          & 0.7831 ± 0.0000          & 0.8501 ± 0.0000      & 0.8204 ± 0.0000     \\
Personal TOP & 0.7545 ± 0.0000          & 0.7701 ± 0.0000          & 0.7901 ± 0.0000     & 0.8067 ± 0.0000    \\
LSTM         & 0.9026 ± 0.0039          & 0.8752 ± 0.0041          & 0.9395 ± 0.0015     & 0.9336 ± 0.0015    \\
GRU          & 0.9119 ± 0.0059          & 0.8860 ± 0.0062          & 0.9418 ± 0.0016     & 0.9357 ± 0.0018    \\
TGN          & {\ul 0.9390 ± 0.0032}    & {\ul 0.9215 ± 0.0040}    & {\ul 0.9430 ± 0.0017}     & {\ul 0.9365 ± 0.0023}    \\
GraphMixer   & 0.8565 ± 0.0078          & 0.8281 ± 0.0079          & 0.7591 ± 0.0015     & 0.7716 ± 0.0018    \\
TGAT         & 0.7094 ± 0.0095          & 0.6410 ± 0.0116          & 0.6563 ± 0.0107     & 0.6420 ± 0.0180    \\
DyGPP        & \textbf{0.9699 ± 0.0008} & \textbf{0.9591 ± 0.0015} & \textbf{0.9697 ± 0.0002}           & \textbf{0.9647 ± 0.0002}          \\
\bottomrule
\end{tabularx}
\end{table}

The DyGPP model achieved the highest accuracy and the best AUC score among all models. This indicates that dynamic graph models can efficiently handle passenger behavior prediction tasks using dynamic graph structures, and these structures can be effectively extended to other sequential models. Other key observations from the table are as follows:

(i) Statistical methods (i.e., TOP and personal TOP), performed poorly in comparison. The statistical methods only considered the historical boarding and alighting counts of passengers, without analyzing specific times, and they did not account for the temporal sequence and potential changes in passenger travel. Compared to the personal TOP method, the TOP method performed better because most individuals had fewer than $100$ interactions, making it difficult to effectively represent individual travel patterns given the significantly larger number of stations compared to the number of interactions.

(ii) The sequential methods (i.e., LSTM and GRU) performed well in the prediction task. These methods predicted future passenger behavior by combining boarding and alighting patterns, fully learning the travel habits of passengers. However, the sequential methods did not consider the impact of station conditions on passenger travel choices and had insufficient predictive ability for travel behavior fluctuations caused by small-scale special circumstances.

(iii) The dynamic graph methods considered the relationships between nodes, thereby accounting for the mutual influence between stations and passengers, which enhanced the prediction accuracy. The TGAT method used interaction occurrence times as an important basis for learning, but due to the limited amount of available data and the coexistence of long-term periodic patterns and short-term abrupt events in passenger travel behavior, TGAT's learning capability was insufficient. The TGN transmitted interaction information between nodes through message aggregation and stored it in the node's memory, effectively learning the features of passengers and stations, thus performing the best among the baselines. However, the dynamic graph methods did not effectively utilize the historical interaction information between passengers and stations, and their temporal characteristics were lost due to the constraints of the graph structure during message passing.

(iv) The DyGPP model performed the best among all the results. DyGPP used serialized historical interaction information to capture the temporal characteristics of user travel habits and employed learnable node features to depict the attributes of passengers and stations. Building on this foundation, DyGPP integrated a time encoder to incorporate the temporal features of interactions, further enhancing the accuracy of the passenger behavior prediction task.

\subsection{Ablation Study}
An ablation study validated the effectiveness of various parts of the model by systematically removing specific modules and analyzing the impact on the overall performance~\cite{meyes2019ablation}. We designed ablation experiments to further verify the effectiveness of each module in DyGPP. This experiment included examining the effectiveness of the edge encoder (eE), the time encoder (tE), the neighbor co-occurrence encoder (coE), the self-attention part of the encoder (cosE), and the cross-attention part (cocE). We removed each module separately, and refer to these as w/o eE, w/o tE, w/o coE, w/o cosE, and w/o cocE. We evaluated the prediction performance of the different variants on the BJSubway-40K dataset.

Table ~\ref{tab_ablation_result1} shows that DyGPP exhibited the best performance. The following conclusions were confirmed in the experiments: (1) Temporal information played a significant role in the passenger behavior prediction task, as evidenced by the comparison with the w/o tE model. This demonstrated DyGPP's ability to track the temporal information of passenger boarding and alighting. (2) The boarding and alighting attributes of passengers proved to be important, which aligns with common sense, as a passenger's boarding behavior is always paired with their alighting behavior. (3) The co-occurrence module we designed also showed a significant impact on performance. The self-attention part of the co-occurrence module represents the passenger's preference for different stations over a period of time, while the cross-attention part directly represents the association between passengers and the stations they attempt to interact with.

\begin{table}[H]
\caption{Effects of different components in DyGPP.}
\label{tab_ablation_result1}
\begin{tabularx}{\textwidth}{YYY}
\toprule
\textbf{Methods}     & \textbf{AP}                       & \textbf{AUC}\\
\midrule
w/o eE      & 0.9681 ± 0.0005          & 0.9562 ± 0.0005\\
w/o tE      & 0.9685 ± 0.0010          & 0.9570 ± 0.0016\\
w/o coE     & 0.8877 ± 0.0072          & 0.8601 ± 0.0085\\
w/o cosE    & 0.9672 ± 0.0014          & 0.9543 ± 0.0022\\
w/o cocE    & 0.8824 ± 0.0114          & 0.8533 ± 0.0135\\
\textbf{DyGPP 
}       & \textbf{0.9699 ± 0.0008} & \textbf{0.9591 ± 0.0015}\\
\bottomrule
\end{tabularx}
\end{table}

\subsection{Temporal Encoder Analysis}
We further conducted experiments to verify the effectiveness of the MLP temporal encoder, and we replaced this encoder with Transformer, LSTM, and Mamba, which are common sequence modeling methods. We evaluated the runtime of the different model variants on the BJSubway-40K dataset, and the results are shown in Table ~\ref{tab_ablation_result2}. The time represents the total time taken for the model to run (including training, validation, and testing).

In the tests for the temporal information encoder, Mamba included a more complex structure and more parameters; LSTM required step-by-step calculations when processing sequence data, making it difficult to parallelize; Transformer achieved more efficient parallel computation through the self-attention mechanism but had higher theoretical complexity; while MLP, with its simple structure, allowed for fully parallel computation. The AP and AUC results show that MLP performed better than the other encoders, demonstrating the effectiveness of using MLP for temporal learning.

\begin{table}[H]
\caption{Performance on different temporal encoders.}
\label{tab_ablation_result2}
\begin{tabularx}{\textwidth}{YYYY}
\toprule
\textbf{Methods}     & \textbf{AP}                       & \textbf{AUC}                      & \textbf{Time Cost (s) }   \\
\midrule
Transformer & 0.9692 ± 0.0006          & 0.9578 ± 0.0010          & 4707.03          \\
LSTM        & 0.9693 ± 0.0009          & 0.9580 ± 0.0018          & 4008.01          \\
DyGPP-mamba & 0.9690 ± 0.0010          & 0.9581 ± 0.0015          & 5641.16          \\
\textbf{DyGPP}       & \textbf{0.9699 ± 0.0008} & \textbf{0.9591 ± 0.0015} & \textbf{3604.43} \\
\bottomrule
\end{tabularx}
\end{table}

\subsection{Parameter Sensitivity}
\begin{figure}[H]
\includegraphics[width=7.5 cm]{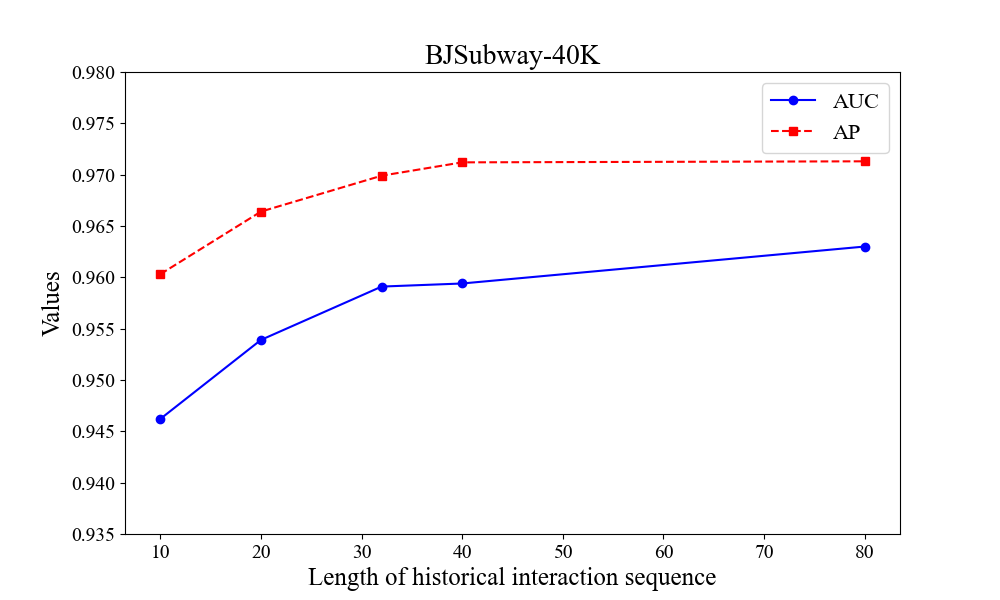}
\caption{Sequence length sensitivity in predicting.}
\label{fig5}
\end{figure}

The historical interaction sequences included the attributes of the nodes themselves and the related information from their neighbors. These sequences had a significant impact on the subgraph construction and node feature representation. Therefore, the length of the interaction sequence was a key factor in determining the prediction accuracy. We tested the parameter sensitivity of DyGPP's prediction results under different sequence lengths in the BJSubway-40K dataset, and the results are depicted in Figure  \ref{fig5}. As seen from the curve, the accuracy tended to improve with increasing sequence length and gradually stabilized. This is because longer sequences provided more information for tracking passenger and station interactions. However, excessively long interaction sequences exceeded the maximum interaction value for passengers, resulting in no further increase in accuracy once a threshold had been reached.

\vspace{-9pt}


\section{Discussion}




In this study, we used the DyGPP model to conduct an in-depth analysis of passenger travel behavior patterns. The results showed that DyGPP outperformed all existing methods in predicting the next station for passengers. This outcome indicates that comprehensively considering the historical interaction information between passengers and stations is crucial in PP tasks. The DyGPP model significantly improved the prediction accuracy of PP tasks, while maintaining a relatively simple architecture and achieving efficient predictions in a short time. This lays a solid foundation for the model's application in real-world production.

Experimental analysis revealed that the neighbor co-occurrence encoder significantly impacted the model's prediction accuracy, with the cross-attention part playing a critical role. Since the model uses first-order neighbors as the basis for sequence construction, and the first-order neighbors of a station are typically passengers, a small amount of non-repetitive passenger information may be insufficient to fully represent the characteristics of a station when the sequence length is short. Additionally, although there are often spatial and logical relationships between stations, this study treated stations as independent nodes for prediction and judgment. This simplification may have affected the prediction accuracy.

Given the limitations of the experiments, future research should focus on two main aspects: first, for station feature extraction, future studies should consider using second-order neighbors to represent stations. Since the second-order neighbors of a station are still stations, this approach can better capture the logical similarities between stations. Second, the integration of spatial location information (including the geographical location of stations and the public transportation routes passing through them) should be considered to represent station features. By doing so, the model can better learn the implicit relationships between stations during training, further improving the prediction accuracy.

The model proposed in this study not only addresses the current issue of traffic management departments relying solely on station flow rate and volume for predictions but also enhances data utilization efficiency by incorporating individual passenger characteristics to more accurately predict future station passenger flow. This method can identify differences among various passenger groups, which can help improve a station's early warning and risk response capabilities in emergencies or special incidents.

\section{Conclusions}
This paper proposed a passenger behavior prediction method based on dynamic graph representation learning. Our approach considers both the long-term and short-term patterns of passenger behavior, defining the passenger behavior prediction problem as a connection prediction problem between heterogeneous nodes in a continuous-time dynamic graph. We modeled the passenger's historical interaction patterns using fixed-length sequences, ensuring robust performance in the face of continuously growing interaction records in the real world. To capture the different behaviors over long and short periods, we employed a sequential representation method to model historical interaction records and used an MLP model to learn passenger behavior characteristics. Finally, to validate the effectiveness of our method, we conducted statistical analyses on datasets and experiments on the Beijing subway dataset. The experiments not only demonstrated the excellent performance of DyGPP but also highlighted its effective behavior prediction capability within dynamic graph models.

\vspace{6pt}

\authorcontributions{Conceptualization, Mingxuan Xie; Methodology, Mingxuan Xie; Validation, Mingxuan Xie; Formal analysis, Mingxuan Xie; Investigation, Mingxuan
Xie; Resources, Mingxuan Xie; Data curation, Mingxuan Xie; Writing - original draft, Mingxuan Xie; Writing – review \& editing, Mingxuan Xie, Tao Zou, Junchen Ye and Bowen Du; Supervision, Bowen Du and Runhe Huang; Funding acquisition, Bowen Du.}

\funding{This research was funded by National Natural Science Foundation of China grant number 51991395, 51991391, U1811463 and the S\&T Program of Hebei grant number 225A0802D.}

  \dataavailability{The data presented in this study are available on request from the corresponding author due to authorization requirements from the data source.}





  \conflictsofinterest{The authors declare no conflicts of interest.}
\begin{adjustwidth}{-\extralength}{0cm} 				
 			
 			\reftitle{References}            

	\PublishersNote{}
\end{adjustwidth}				

\end{document}